\documentclass{article}
\usepackage{multirow}
\usepackage{spconf,amsmath,graphicx,hyperref}
\usepackage[absolute,overlay]{textpos}
\usepackage{booktabs}
\usepackage{amssymb}

\title{DARC-CLIP: Dynamic Adaptive Refinement with Cross-Attention for Meme Understanding}
\name{Qiyuan Jin}
\address{The Hong Kong University of Science and Technology}
\begin{document}
\ninept
\begin{textblock*}{0.825\paperwidth}(1.85cm,26.3cm)
\noindent
\footnotesize
\textcopyright~2026 IEEE. Personal use of this material is permitted. Permission from IEEE must be obtained for all other uses, in any current or future media, including reprinting/republishing this material for advertising or promotional purposes, creating new collective works, for resale or redistribution to servers or lists, or reuse of any copyrighted component of this work in other works. DOI: 10.1109/ICASSP55912.2026.11462868
\end{textblock*}
\maketitle

\begin{abstract}
Memes convey meaning through the interaction of visual and textual signals, often combining humor, irony, and offense in subtle ways. Detecting harmful or sensitive content in memes requires accurate modeling of these multimodal cues. Existing CLIP-based approaches rely on static fusion, which struggles to capture fine-grained dependencies between modalities. We propose DARC-CLIP, a CLIP-based framework for adaptive multimodal fusion with a hierarchical refinement stack. DARC-CLIP introduces Adaptive Cross-Attention Refiners to for bidirectional information alignment and Dynamic Feature Adapters for task-sensitive signal adaptation. We evaluate DARC-CLIP on the PrideMM benchmark, which includes hate, target, stance, and humor classification, and further test generalization on the CrisisHateMM dataset. 
DARC-CLIP achieves highly competitive classification accuracy across tasks, with significant gains of +4.18 AUROC and +6.84 F1 in hate detection over the strongest baseline. Ablation studies confirm that ACAR and DFA are the main contributors to these gains. These results show that adaptive cross-signal refinement is an effective strategy for multimodal content analysis in socially sensitive classification. 

\end{abstract}
\section{Introduction}
\label{sec:intro}
The development of social media has made text-embedded images, known as memes, a dominant form of online communication~\cite{doi:10.1177/1461444820912722}. While memes serve as a powerful tool for sharing opinions, they are also increasingly used to covertly spread hate speech and misinformation, posing serious threats to a healthy online ecosystem \cite{shah2024memeclipleveragingcliprepresentations}. Meme interpretation is difficult: an image signal that appears harmless in isolation may become derogatory when combined with specific text. Cultural context further complicates reliable classification, especially in sensitive domains such as LGBTQ+ movement~\cite{imperato2023all,griffin}. Effective meme understanding thus requires multimodal reasoning across multiple aspects including hate, stance, humor, and target identification~\cite{kiela2021hateful}.

Early hate speech detection primarily focused on text-only analysis~\cite{alam2022survey,10.1007/s00530-023-01051-8,10461793}, overlooking the interplay between visual and textual cues in memes. Several architectures have been proposed to address these challenges. MOMENTA~\cite{pramanick2021momentamultimodalframeworkdetecting}  uses cross-modality attention fusion
to further modulate the concatenated features.
CLIP~\cite{radford2021learningtransferablevisualmodels} introduced a unified semantic space for images and text signals without task-specific pretraining, inspiring multiple extensions. 
HateCLIPper~\cite{kumar2022hateclippermultimodalhatefulmeme} employs bilinear pooling to capture fine-grained interactions. MemeCLIP~\cite{shah2024memeclipleveragingcliprepresentations} improves parameter efficiency with lightweight feature adapters and semantic-aware initialization in multi-aspect classification tasks.
Dataset growth has also driven progress. HMC~\cite{kiela2021hateful}, HarMeme~\cite{pramanick2021momentamultimodalframeworkdetecting}, DisinfoMeme~\cite{qu2022disinfomememultimodaldatasetdetecting}, CrisisHateMM~\cite{Bhandari_2023_CVPR}, and PrideMM~\cite{shah2024memeclipleveragingcliprepresentations} introduce domain-specific annotations, including multi-aspect labels in real-world memes.

Despite these advances, most existing approaches rely on static fusion mechanisms or minimally adapted CLIP features, limiting their ability to model dynamic, context-dependent modality interactions. These constraints hinder generalization under domain shifts and multi-aspect classification tasks, which motivates the need for our proposed dynamic and adaptive fusion approach.

We address this problem with DARC-CLIP, an efficient CLIP-based framework for adaptive multimodal fusion. DARC-CLIP introduces two modules: Adaptive Cross Attention Refiners (ACAR) that iteratively align modality-specific features, and Dynamic Feature Adapters (DFA) that recalibrate feature distributions across tasks. This hierarchical design of progressive refinement stack enables deeper cross-modal interactions while preserving the generalization benefits of pre-trained models. Lightweight linear projections, a cosine classifier with semantic-aware initialization, and optional partial CLIP unfreezing further improve adaptability.

Our contributions are summarized as follows:
(1) A hierarchical multimodal refinement stack combining ACAR and DFA for fine-grained, context-aware multimodal reasoning;
(2) Strong performance on 
PrideMM~\cite{shah2024memeclipleveragingcliprepresentations} and CrisisHateMM~\cite{Bhandari_2023_CVPR}, demonstrating cross-domain robustness;
(3) Comprehensive ablations showing the complementary benefits of our core components and provide practical insights into the accuracy–efficiency trade-offs.

\section{Methodology}
\label{sec:method}
In this section, we present DARC-CLIP, a framework for multimodal meme classification. It augments a pre-trained CLIP backbone with a novel hierarchical refinement stack to enable deeper, more flexible cross-modal interactions. 

\begin{figure*}
    \centering
    \includegraphics[width=\textwidth]{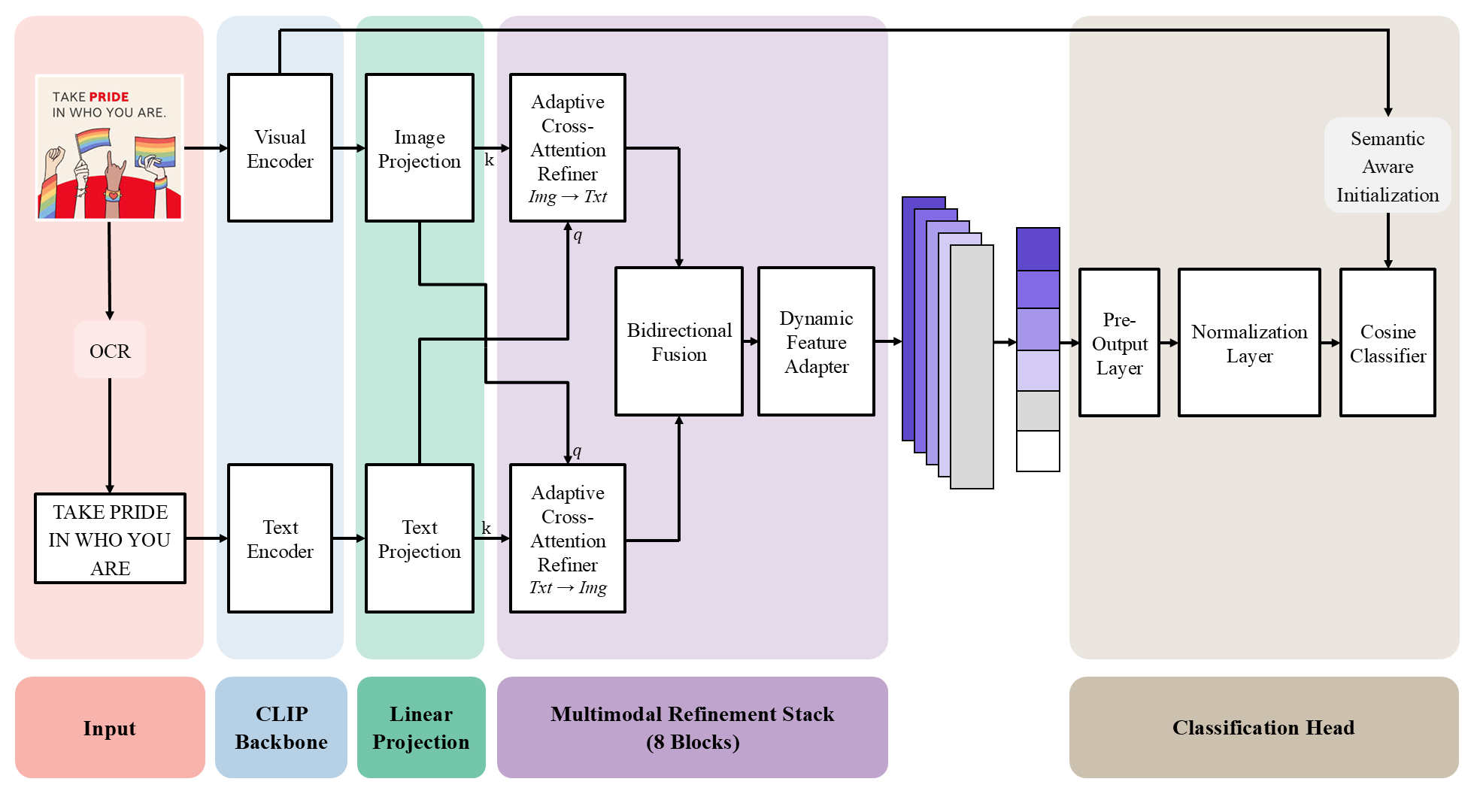}
    \caption{An overview of our proposed framework, DARC-CLIP.}
    \label{fig:arch}
\end{figure*}

DARC-CLIP begins with CLIP~\cite{radford2021learningtransferablevisualmodels} for initial visual and textual representations. To better handle domain-specific semantics, selected layers of CLIP are partially unfrozen for controlled fine-tuning. The raw features from these encoders are linearly projected into a shared embedding space \cite{shah2024memeclipleveragingcliprepresentations, clippae} and processed by a stack of refinement blocks, named as Multimodal Refinement Stack. Each block  progressively integrates cross-signal information through two lightweight modules: an Adaptive Cross-Attention Refiner (ACAR) and a Dynamic Feature Adapter (DFA). Finally, we aggregate outputs across blocks and classify the resulting feature with a cosine classifier initialized from CLIP’s text prototypes. The overall pipeline is shown in Fig.~\ref{fig:arch}.

\subsection{Adaptive Cross-Attention Refiner (ACAR)}

The design of ACAR employs bidirectional refinement specifically tailored to capture the subtle semantic contradictions in memes, such as textual irony that hinges on specific visual context. We first map raw features $\mathbf{F}_{\text{img}}, \mathbf{F}_{\text{txt}}\!\in\!\mathbb{R}^{768}$ from CLIP's encoders into a higher-dimensional space $\mathbb{R}^{D_\text{map}}(D_\text{map}=1024)$ via linear projections:
\begin{equation}
\mathbf{F}'_{\text{mod}} = \text{ReLU}(\mathbf{W}_{\text{mod}} \mathbf{F}_{\text{mod}} + \mathbf{b}_{\text{mod}})
\label{eq:proj}
\end{equation}
These projected signals are fed into $L$ refinement blocks. In each block $l$, $\text{ACAR}^{(l)}(\mathbf{Q}, \mathbf{KV})$ refines the query $\mathbf{Q}$ by attending to key-value pairs derived from $\mathbf{K}$. Specifically, a multi-head cross-attention mechanism first captures cross-modal context\cite{chen2024,vaswani2023attentionneed}:
\begin{equation}
\begin{aligned}
    \mathbf{A}^{(l)} = \text{Concat}(\text{head}_1^{(l)},\dots,\text{head}_H^{(l)})\mathbf{W}^{O^{(l)}}\ \ \ \ \ \ 
\\[2pt]
    \text{head}_h^{(l)} = \text{softmax}\!\left(\frac{\mathbf{Q}\mathbf{W}_Q^{(h,l)}(\mathbf{K}\mathbf{W}_K^{(h,l)})^\top}{\sqrt{d_k}}\right)\mathbf{V}\mathbf{W}_V^{(h,l)}
\end{aligned}
\end{equation}

To improve expressiveness with minimal cost, ACAR includes an adaptive MLP with learnable scaling \cite{dong2025,10138903}:
\begin{equation}
\widetilde{\mathbf{A}}^{(l)}=\lambda^{(l)}\,\mathbf{W}_\uparrow^{(l)}\,
\text{GELU}(\mathbf{W}_\downarrow^{(l)}\mathbf{A}^{(l)}),
\label{eq:mlp}
\end{equation}
where $\lambda^{(l)}$ modulates the contribution of this branch. Specifically, $\lambda^{(l)}$ is initialized to 0.05 to ensure a stable warm-up of the adaptive branch during the initial stages of training. ACAR aggregates the original query, the attention output, and the adaptive branch via residual integration and layer normalization:
\begin{equation}
\operatorname{ACAR}^{(l)}(\mathbf{Q},\mathbf{K}\mathbf{V})=
\operatorname{LN}(\mathbf{Q}+\mathbf{A}^{(l)}+\widetilde{\mathbf{A}}^{(l)}).
\label{eq:acar}
\end{equation}
ACAR is applied in both directions $\mathbf{Z}^{(l)}_{\text{img}\rightarrow\text{txt}} = \text{ACAR}^{(l)}(\mathbf{F}'_{\text{img}}, \mathbf{F}'_{\text{txt}})$ and $\mathbf{Z}^{(l)}_{\text{txt}\rightarrow\text{img}} = \text{ACAR}^{(l)}(\mathbf{F}'_{\text{txt}}, \mathbf{F}'_{\text{img}})$ to support bidirectional cross-signal alignment. This design ensures stability, preserves original semantics, and integrates adaptively transformed information for robust multimodal alignment.

\subsection{Dynamic Feature Adapter (DFA)}
While ACAR facilitates cross-modal interaction, additional refinement is required to capture task-specific nuances and avoid negative transfer in multi-task scenarios. To address this, we propose the Dynamic Feature Adapter (DFA) with gating-based adaptive mechanism\cite{dfsq,kawano2023}. DFA dynamically regulates its influence based on global context, enabling selective feature and task adaptation where necessary.

DFA operates on the fused representation from the ACAR outputs in block $l$:
\begin{equation}
\mathbf{Z}^{(l)}_{\text{fuse}}=\tfrac{1}{2}\big(
\mathbf{Z}^{(l)}_{\text{img}\rightarrow\text{txt}}+
\mathbf{Z}^{(l)}_{\text{txt}\rightarrow\text{img}}\big).
\label{eq:fuse}
\end{equation}
It first computes a gating score that reflects the input’s global semantics, passed through a linear layer and sigmoid:
\begin{equation}
\mathbf{g}^{(l)} = \text{Sigmoid}(\mathbf{W}^{(l)}_g \cdot \text{Mean}(\mathbf{Z}^{(l)}_{\text{fuse}}) + b^{(l)}_g)
\label{eq:gate}
\end{equation}
This gating mechanism adaptively modulates the strength of feature transformation for each sample, allowing task-sensitive control. 
To refine $\mathbf{Z}^{(l)}_{\text{fuse}}$, DFA employs a lightweight bottleneck MLP similar to ACAR but scaled by the dynamic gate:
\begin{equation}
\mathbf{H}^{(l)}_{\text{gated}} = \mathbf{g}^{(l)} \odot \big( \text{MLP}^{(l)}(\mathbf{Z}^{(l)}_{\text{fuse}}) \big)
\end{equation}
The final representation is obtained via a residual connection and layer normalization, ensuring stability while achieving dynamic, task-sensitive adaptation.
\begin{equation}
\operatorname{DFA}^{(l)}=\operatorname{LN}(\mathbf{Z}^{(l)}_{\text{fuse}}+\mathbf{H}^{(l)}).
\label{eq:dfa}
\end{equation}
This mechanism selectively adapts features based on global context through dynamic gating, improving robustness under task and domain variability and ensuring stability during optimization via residual normalization.

\subsection{Hierarchical Feature Aggregation and Classification}
To capture multi-level abstractions, we aggregate the outputs from all $L$ refinement blocks through uniform averaging \cite{wu2023bid,zhou2023att}:
\begin{equation}
\mathbf{F}_{\text{final}} = \frac{1}{L} \sum_{l=1}^{L} \text{DFA}^{(l)}(\mathbf{Z}_{\text{fuse}}^{(l)})
\label{eq:agg}
\end{equation}
This hierarchical aggregation combines coarse and fine-grained cues across blocks, resulting in a more comprehensive multimodal representation compared to single-block fusion.

The final aggregated feature $\mathbf{F}_{\text{final}}$ is classified using a cosine classifier \cite{liu2020deep}. 
\begin{equation}
    Z_c = \sigma \cdot \frac{\mathbf{W}_c \cdot \mathbf{F}_{\text{final}}}{\|\mathbf{W}_c\|_2 \|\mathbf{F}_{\text{final}}\|_2}
\end{equation}
where $\mathbf{W}_c$ is initialized from CLIP text embeddings corresponding to natural-language prompts for class $c$, and $\sigma$ is a fixed scaling factor. This initialization aligns classifier prototypes with textual semantics, improving calibration and generalization at negligible computational  cost\cite{shah2024memeclipleveragingcliprepresentations}.

\section{Experiments}
\label{sec:exp}
This section presents a comprehensive evaluation of DARC-CLIP, demonstrating its effectiveness and generalization across domains. We first detail our experimental setup, followed by a comparison against strong baselines and an ablation study to validate the contribution of each component.


\subsection{Experimental Setup}

We evaluate DARC-CLIP on two meme benchmarks: the PrideMM \cite{shah2024memeclipleveragingcliprepresentations} and the CrisisHateMM \cite{Bhandari_2023_CVPR}. PrideMM contains 5,063 memes related to the LGBTQ+ Pride movement with multi-aspect annotations for four tasks: Hate Speech Detection, Target Classification, Topical Stance Classification, and Intended Humor Detection. 
It exhibits significant class imbalance, as summarized in Table~\ref{tab:dataset}, posing challenging for robust multimodal reasoning. CrisisHateMM provides over 4,700 memes 
for hate speech detection, and is used to assess generalization under domain shift. For PrideMM, we follow the official splits and further divide the training set (90\%/10\%) for training and internal validation. All reported results use the official validation set. CrisisHateMM is used solely for cross-domain evaluation. We report Accuracy, Macro F1 Score, and AUROC, with F1 and AUROC being particularly crucial for class-imbalanced tasks.

We use ViT-L/14 CLIP as the backbone. Models were trained for 15 epochs with a batch size of 32 on NVIDIA RTX 4090 GPU. The best checkpoint is selected based on validation AUROC. Each experiment was repeated three times; the mean scores are reported.

\begin{table}
\centering
  \caption{Label distribution for the publicly available subset of the PrideMM dataset (train + validation splits combined).}
  \label{tab:dataset}
  \begin{tabular}{c c c c}
    \toprule
    \textbf{Task} & \textbf{Label} & \textbf{\#Samples} & \textbf{\%} \\
    \midrule
    \multirow{2}{*}{\centering Hate}
    & No Hate & 2,313 & 49.23\% \\ 
    & Hate & 2,243 & 50.77\% \\  
    \midrule
    \multirow{2}{*}{\centering Target}
    & Undirected & 694 & 31.08\% \\
    & Individual & 224 & 10.03\% \\ 
    & Community & 1,047 & 46.89\% \\ 
    & Organization & 268 & 12.00\% \\ 
    \midrule
    \multirow{2}{*}{\centering Stance}
    & Neutral & 1,312 & 28.80\% \\ 
    & Support & 1,718 & 37.71\% \\ 
    & Oppose & 1,526 & 33.49\% \\ 
    \midrule
    \multirow{2}{*}{\centering Humor}
    & No Humor & 1,477 & 31.72\% \\ 
    & Humor & 3,179 & 68.28\% \\ 
  \bottomrule
\end{tabular}
\end{table}

\subsection{Baseline Comparison}
We compare DARC-CLIP against a range of strong unimodal (BERT \cite{devlin2019bertpretrainingdeepbidirectional}, ViT-L/14 \cite{dosovitskiy2021imageworth16x16words}) and multimodal baselines (CLIP \cite{radford2021learningtransferablevisualmodels}, HateCLIPper \cite{kumar2022hateclippermultimodalhatefulmeme}, ISSUES \cite{burbi2023mappingmemeswordsmultimodal}, MemeCLIP \cite{shah2024memeclipleveragingcliprepresentations}). 

\begin{figure}
  \centering
  \includegraphics[width=0.6\linewidth]{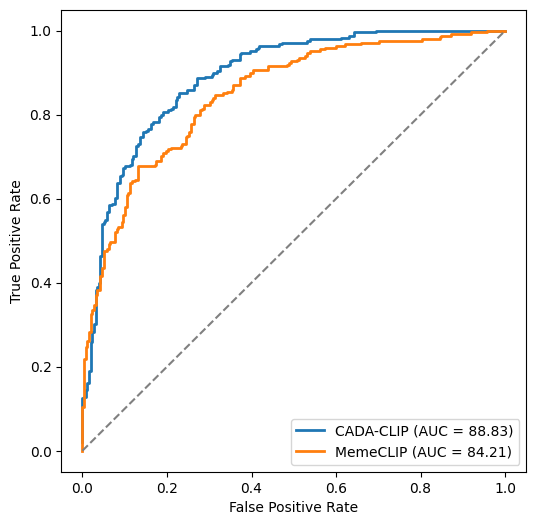}
  \caption{ROC curves of MemeCLIP and DARC-CLIP for Hate Speech Detection on PrideMM.}
  \label{fig:roc}
\end{figure}

Table~\ref{tab:performonpride} shows the results on PrideMM dataset. Unimodal methods perform substantially worse than multimodal ones, highlighting the importance of jointly modeling textual and visual signals in memes. DARC-CLIP consistently delivers the best overall results across four tasks, outperforming the best baseline MemeCLIP. Substantial F1 score improvements (e.g., +6.8 on Hate) further validate our approach. The superior capability is further demonstrated by ROC curves in Fig.~\ref{fig:roc} for Hate Speech Detection, where DARC-CLIP outperforms MemeCLIP across thresholds. 
To further assess robustness to domain shift, we evaluate our model on the CrisisHateMM dataset. As shown in Table~\ref{tab:performcrisis}, DARC-CLIP achieves the highest AUROC. This result confirms that our framework not only excels on the target domain but also exhibits strong cross-domain generalization, making it highly applicable to diverse real-world scenarios.

\begin{table*}
  \centering
  \caption{Performance metrics for different methods on the PrideMM dataset.}
  \label{tab:performonpride}
  \begin{tabular}{*{13}{c}}
    \toprule
    \multirow{2}{*}{\textbf{Method}} & \multicolumn{3}{c}{\textbf{Hate}} & \multicolumn{3}{c}{\textbf{Target}} & \multicolumn{3}{c}{\textbf{Stance}} & \multicolumn{3}{c}{\textbf{Humor}} \\
    \cmidrule(lr){2-4} \cmidrule(lr){5-7} \cmidrule(lr){8-10} \cmidrule(lr){11-13}
    & \textbf{Acc.} & \textbf{AUC.} & \textbf{F1} & \textbf{Acc.} & \textbf{AUC.} & \textbf{F1} & \textbf{Acc.} & \textbf{AUC.} & \textbf{F1} & \textbf{Acc.} & \textbf{AUC.} & \textbf{F1} \\
    \midrule
    BERT & 71.23 & 75.82 & 71.15 & 51.42 & 76.36 & 51.03 & 53.10 & 70.75 & 51.48 & 68.93 & 75.21 & 67.63 \\
    ViT-L/14 & 50.83 & 64.51 & 59.15 & 53.58 & 69.83 & 51.29 & 51.94 & 66.49 & 49.61 & 59.26 & 70.53 & 65.43 \\
    \midrule
    CLIP & 72.78 & 79.27 & 73.06 & 58.82 & 78.92 & 55.37 & 57.35 & 73.38 & 56.63 & 73.29 & 79.43 & 72.04 \\
    HateCLIPper & 73.46 & 78.43 & 71.62 & 56.34 & 73.81 & 51.26 & \textbf{63.19} & 71.24 & 55.69 & 71.43 & 77.86 & 69.24 \\
    ISSUES & 75.38 & 83.27 & 72.54 & 60.34 & 79.82 & 54.24 & 62.12 & 77.34 & 60.49 & 76.51 & 82.13 & 70.43 \\
    MemeCLIP & 74.09 & 84.45 & 73.68 & \textbf{60.89} & 77.31 & 53.03 & 62.83 & 79.26 & 61.17 & 79.64 & 82.88 & 71.25 \\
    \midrule
    DARC-CLIP & \textbf{80.84} & \textbf{88.63} & \textbf{80.52} & 60.45 & \textbf{82.58} & \textbf{57.88} & 62.26 & \textbf{80.97} & \textbf{61.78} & \textbf{80.13} & \textbf{83.45} & \textbf{76.10} \\
    \bottomrule
  \end{tabular}
\end{table*}

\begin{table}
  \centering
  \caption{Performance metrics for different methods on the CrisisHateMM dataset.}
  \label{tab:performcrisis}
  \begin{tabular}{c@{\hspace{0.5cm}}ccc}
    \toprule
    {\textbf{Method}} & \textbf{Acc.} & \textbf{AUC.} & \textbf{F1} \\
    \midrule
    BERT & 76.65 & 81.26 & 74.68 \\
    ViT-L/14 & 73.42 & 80.46 & 72.97 \\
    \midrule
    CLIP & 79.34 & 86.57 & 78.26 \\
    HateCLIPper & 80.57 & 83.91 & 75.13 \\
    ISSUES & 83.63 & 88.34 & 75.99 \\
    MemeCLIP & 82.39 & 89.62 & 79.41 \\
    \midrule
    DARC-CLIP & \textbf{85.31} & \textbf{92.52} & \textbf{83.28} \\
    \bottomrule
  \end{tabular}
\end{table}

\subsection{Ablation Studies}

We conduct ablation studies on PrideMM to isolate the contribution of each component, with results summarized in Table~\ref{tab:ablation}. Introducing ACAR on top of the frozen CLIP brings a substantial improvement across all metrics, highlighting its key role in enhancing cross-modal alignment and refining feature representations. Adding DFA further improves performance, confirming the synergy of our dual-module refinement stack. In contrast, auxiliary components (SAI, LP) offer only marginal gains; while partial unfreezing provides a slight boost at the cost of 1.3$\times$ longer training time. 
Overall, ACAR and DFA are the dominant contributors to performance improvement, validating the effectiveness of our adapter-based design. Auxiliary strategies such as SAI, LP, and partial unfreezing provide only minor additional benefits.

\begin{table}[t]
\centering
\caption{
Ablation study of DARC-CLIP on the PrideMM hate detection task. Frozen denotes a fully frozen CLIP backbone. Unfrozen indicates partial unfreezing CLIP.
}
\label{tab:ablation}
\resizebox{\columnwidth}{!}
{
\begin{tabular}{cccccccc}
\toprule
\textbf{CLIP} & \textbf{ACAR} & \textbf{DFA} & \textbf{SAI} & \textbf{LP} & \textbf{Acc.} & \textbf{AUC.} & \textbf{F1} \\
\midrule
Frozen    &             &             &             &             & 72.78 & 79.27 & 73.06 \\
Frozen    & \checkmark  &             &             &             & 78.23 & 85.34 & 78.30 \\
Frozen    & \checkmark  & \checkmark  &             &             & 80.14 & 87.83 & 79.51 \\
Frozen    & \checkmark  & \checkmark  & \checkmark  &             & 80.35 & 88.12 & 79.37 \\
Frozen    & \checkmark  & \checkmark  & \checkmark  & \checkmark  & 80.43 & 88.53 & 79.97 \\
\midrule
Unfrozen  & \checkmark  & \checkmark  & \checkmark  & \checkmark  & \textbf{80.84} & \textbf{88.63} & \textbf{80.52} \\
\bottomrule
\end{tabular}
}
\end{table}

\subsection{Qualitative Analysis}

\begin{figure}[t]
\centering
\renewcommand{\arraystretch}{1}
\setlength{\tabcolsep}{8pt}
\begin{tabular}{c@{\hspace{0.5cm}}c}
\begin{minipage}{0.45\linewidth}
    \centering
    \includegraphics[height=2.8cm]{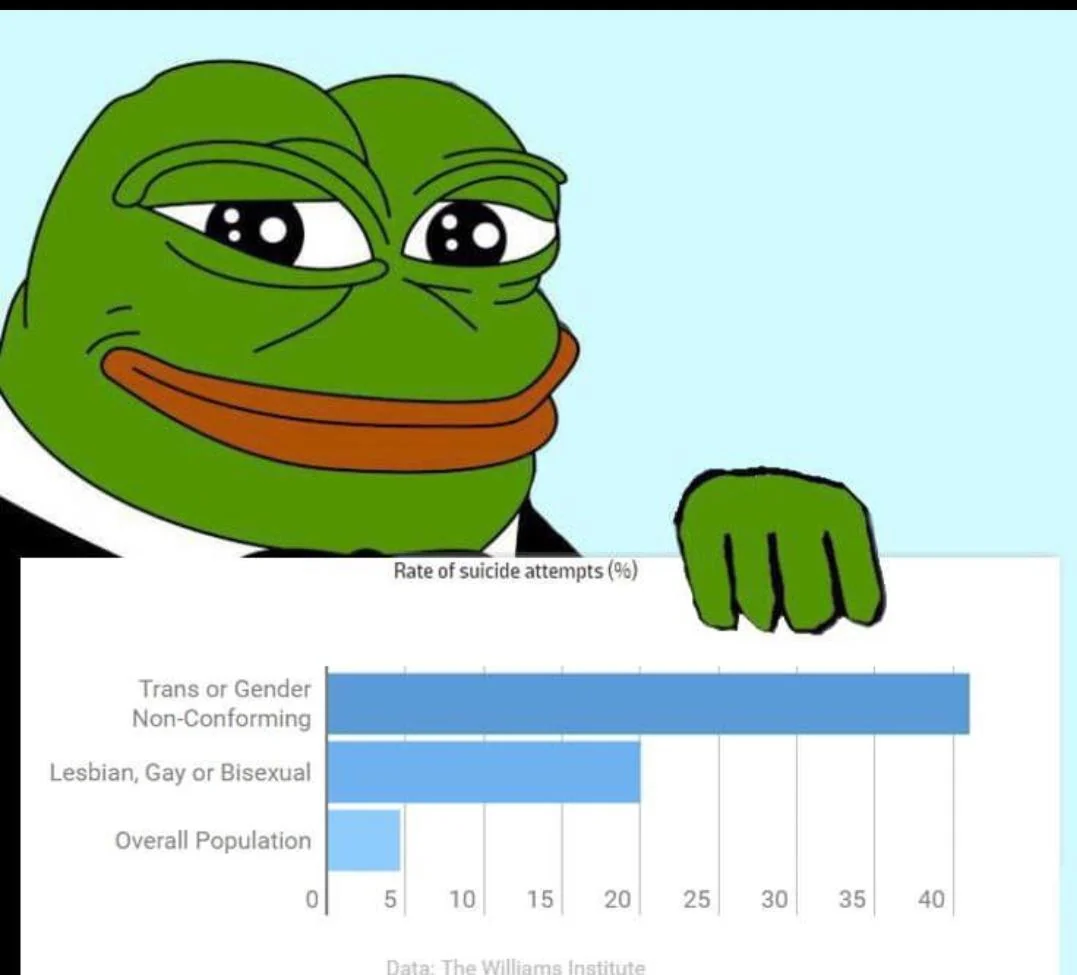} \\
    \small
    (a) \\
    True: \{1, 2, 2, 1\} \\
    MemeCLIP: \{0, 0, 0, 0\} \\
    DARC-CLIP: \{1, 2, 2, 1\} 
\end{minipage}
&
\begin{minipage}{0.45\linewidth}
    \centering
    \includegraphics[height=2.8cm]{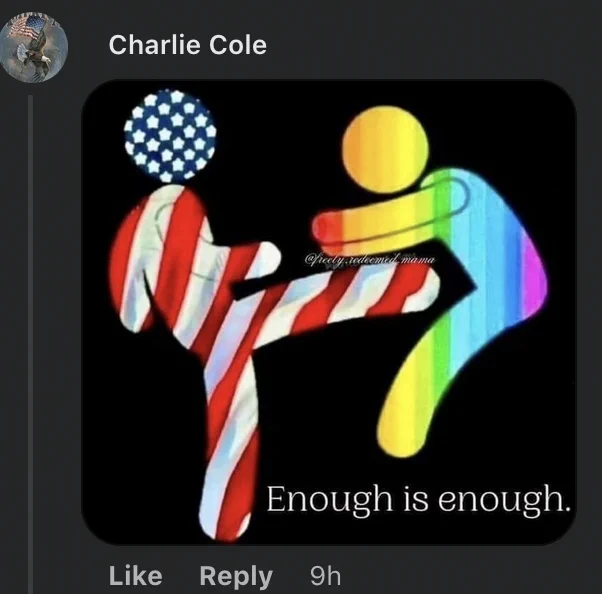} \\
    \small
    (b) \\
    True: \{1, 2, 2, 0\} \\
    MemeCLIP: \{0, 3, 2, 0\} \\
    DARC-CLIP: \{1, 3, 2, 0\} 
\end{minipage}
\end{tabular}
\caption{Qualitative error analysis on PrideMM dataset. Labels correspond to tasks in order: Hate (0=No Hate, 1=Hate), 
Target (0=Undirected, 1=Individual, 2=Community, 3=Organization), 
Stance (0=Neutral, 1=Support, 2=Oppose), 
Humor (0=No Humor, 1=Humor).}

\label{fig:qualitative}
\end{figure}

To better understand the behavior of DARC-CLIP, we present representative cases in Fig.~\ref{fig:qualitative}. 
Case (a) contains implicit visual aggression; DARC-CLIP correctly predicts all labels by integrating textual and visual signals, while MemeCLIP misclassifies them due to its shallow fusion strategy, demonstrating the strength of deep refinement stack. 
Case (b) illustrates a scenario dominated by chart-based information; both models mispredict stance, yet DARC-CLIP achieves more accurate hate detection compared to MemeCLIP. These examples further highlight DARC-CLIP's advantage in capturing implicit multimodal cues compared to single-step fusion. A key limitation is the model's struggle with niche cultural symbols and the difficulty in disentangling image subjects from specific hate speech targets. Consequently, effectively handling complex semantic inference and visual-textual signals integration remain challenges for future work.

\section{Conclusion}
\label{sec:conclusion}
We introduced DARC-CLIP, a novel adaptive multimodal framework for fine-grained meme classification. By introducing Adaptive Cross-Attention Refiners and Dynamic Feature Adapters, our model achieves deeper and more flexible integration of visual and textual signals, addressing the complex semantics in memes. Experiments on the PrideMM and CrisisHateMM demonstrate that DARC-CLIP consistently outperforms prior methods, validating its effectiveness and strong cross-domain generalization ability.
Ablation studies reveal that the majority of performance gains come from ACAR and DFA, while auxiliary strategies Semantic-Aware Initialization and Linear Projection offer marginal complementary benefits. Importantly, DARC-CLIP retains competitive accuracy even when the CLIP backbone remains frozen, highlighting the design's efficiency in resource-constrained scenarios.
Despite these advances, challenges remain in handling highly implicit or abstract content. Our future research aims to incorporate external knowledge bases to better model cultural context in different languages and further enhance the framework's generalization capabilities.


\vfill
\newpage




\bibliographystyle{IEEEtran}
\bibliography{strings,refs}

\end{document}